\documentclass{article}
\usepackage{dcase2022,amsmath,graphicx,url,times,booktabs, tabularx, xcolor}
\usepackage{siunitx}
\usepackage{hyperref,url}
\usepackage{caption,subcaption}

\title{Sound Scene Synthesis at the DCASE 2024 Challenge}

\name{Mathieu Lagrange$^{1}$,
      Junwon Lee$^{2, 3}$,
      Modan Tailleur$^{1}$,
      Laurie M. Heller$^{4}$,
      }
\secondlinename{
      Keunwoo Choi$^{2}$,
      Brian McFee$^{5}$,
      Keisuke Imoto$^{6}$,
      Yuki Okamoto$^{7}$
      }

\address{$^1$ CNRS, Ecole Centrale Nantes, Nantes Université, France, mathieu.lagrange@ls2n.fr\\
         $^2$ Gaudio Lab, Inc., South Korea, \{junwon.lee, keunwoo\}@gaudiolab.com\\
         $^3$ Music and Audio Computing Lab, KAIST, South Korea\\
         $^4$ Carnegie Mellon University, USA, laurieheller@cmu.edu\\
         $^5$ New York University, USA, brian.mcfee@nyu.edu\\
         $^6$ Doshisha University, Japan, keisuke.imoto@ieee.org\\
         $^7$ The University of Tokyo, Japan, y-okamoto@ieee.org\\
}

\begin{document}

\ninept
\maketitle
\begin{sloppy}
\begin{abstract}
This paper presents Task 7 at the DCASE 2024 Challenge: sound scene synthesis. Recent advances in sound synthesis and generative models have enabled the creation of realistic and diverse audio content. We introduce a standardized evaluation framework for comparing different sound scene synthesis systems, incorporating both objective and subjective metrics. The challenge attracted four submissions, which are evaluated using the Fréchet Audio Distance (FAD) and human perceptual ratings. Our analysis reveals significant insights into the current capabilities and limitations of sound scene synthesis systems, while also highlighting areas for future improvement in this rapidly evolving field.
\end{abstract}

\section{Introduction}
This paper presents Task 7 at the DCASE 2024 Challenge: sound scene synthesis. The challenge is motivated by the recent advances generative models for the creation of realistic and diverse audio content, as proposed in \cite{choi2022proposal} and following the last year's version \cite{foley2023}. 

\section{Problem and Task Definition}
We defined the challenge as a text-to-sound generation task, where systems must generate realistic environmental audio based on textual descriptions. This is a more flexible setup than the category-based generation used in the last year \cite{foley2023}.

Each prompt follows the following structure: 

\begin{center}
"\textit{Foreground} with \textit{Background} in the background," 
\end{center}

to specify both the primary sound source and its acoustic context separately.

Key constraints for the generated audio include:
\begin{itemize}
    \item 4-second 16-bit mono audio snippets at 32~kHz sampling rate
    \item No music allowed in the generated audio
    \item No intelligible speech permitted
\end{itemize}

The task emphasizes generative approaches rather than retrieval-based methods, requiring systems to synthesize novel audio rather than simply copy existing samples.

\section{Dataset and Baseline}

\subsection{Dataset Creation}
The challenge dataset contains 310 audio-captions in total, with 60 samples designated for development and 250 for evaluation. All audio content was carefully designed by a sound engineer to match specific prompts, ensuring high-quality and consistent sound scenes. The audio samples were sourced from Freesound.org and the private libraries Rabbit Ears Audio, Sound Ideas, Euro S Phere, The Art Of Foley, BBC Sound Effect, and HissAndARoar, all of which were selected following strict quality guidelines.

\subsection{Sound Categories}
The dataset organized sound content into two main categories: foreground and background sounds. The foreground category encompasses six distinct types of sounds: \textit{Animal, Vehicle, Human, Alarm, Tool, and Entrance} sounds. These were chosen to represent a diverse range of common sound sources in everyday environments. For background sounds, the dataset includes five categories: \textit{Crowd, Traffic, Water, Birds, and Room Tone\footnote{Room tone is a recorded sound with no specific sound event and used to capture natural noise of a recording environment.} (labeled as "\textit{Nothing}")} sounds . This categorization enables the creation of realistic sound scenes with clear foreground-background relationships.

\subsection{Baseline System}
For the baseline implementation, we utilized AudioLDM~\cite{liu2023audioldm} as the core synthesis engine. The system was trained on a comprehensive collection of audio datasets including AudioCaps~\cite{audiocaps}, AudioSet~\cite{audioset}, FreeSound\footnote{\url{https://freesound.org/}}, and BBC Sound Effect datasets\footnote{\url{https://sound-effects.bbcrewind.co.uk/search}}. This diverse training data ensures the baseline system can handle a wide range of sound types and acoustic environments represented in our challenge.

\section{Evaluation Methodology}

\begin{figure*}[th!]
    \centering 
    \def\figwidth{0.2\textwidth}
    \begin{subfigure}{\figwidth}
        \centering
        \includegraphics[width=\linewidth]{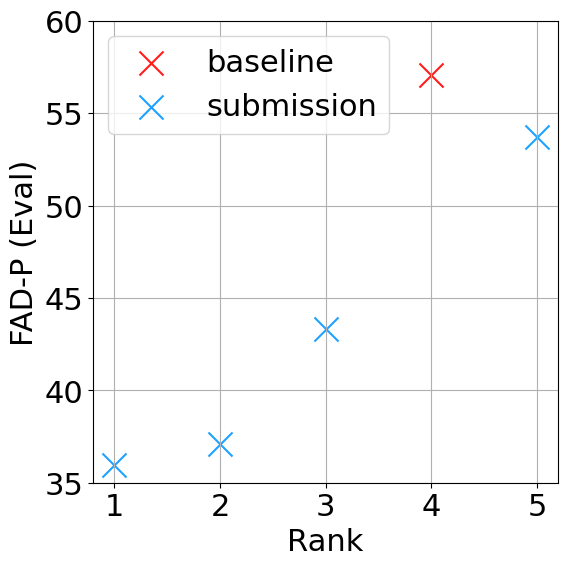}
        \subcaption{FAD-P on Evaluation set vs Challenge Ranking}
        \label{fig:fad-rank}
    \end{subfigure}
    \begin{subfigure}{\figwidth}
        \centering
        \includegraphics[width=\linewidth]{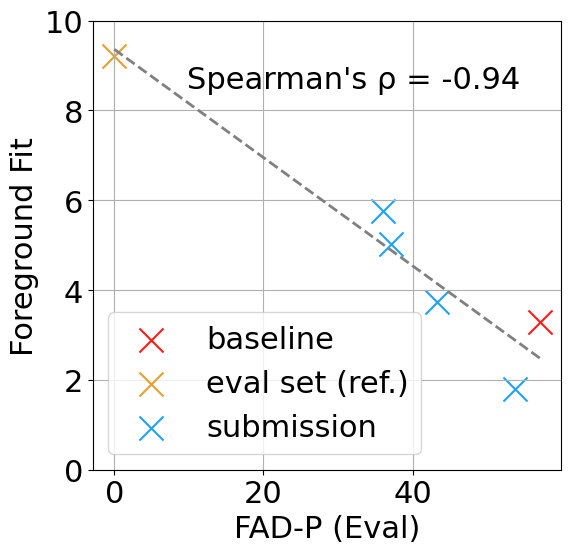}
        \subcaption{FAD-P on Evaluation set vs Foreground Fit}
        \label{fig:eval-mos-fg}
    \end{subfigure}
    \begin{subfigure}{\figwidth}
        \centering
        \includegraphics[width=\linewidth]{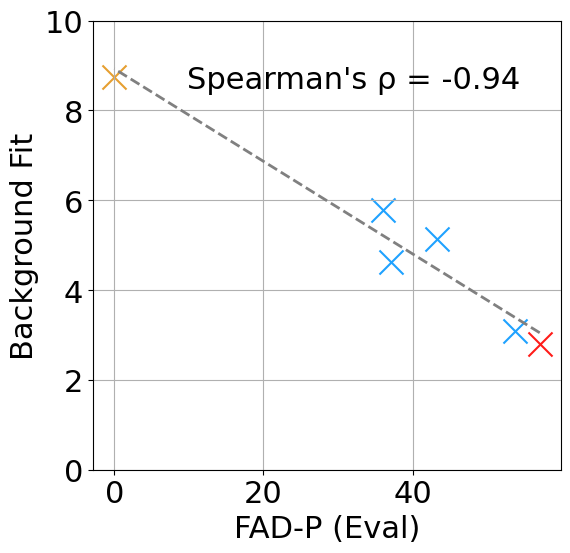}
        \subcaption{FAD-P on Evaluation set vs Background Fit}
        \label{fig:eval-mos-bg}
    \end{subfigure}
    \begin{subfigure}{\figwidth}
        \centering
        \includegraphics[width=\linewidth]{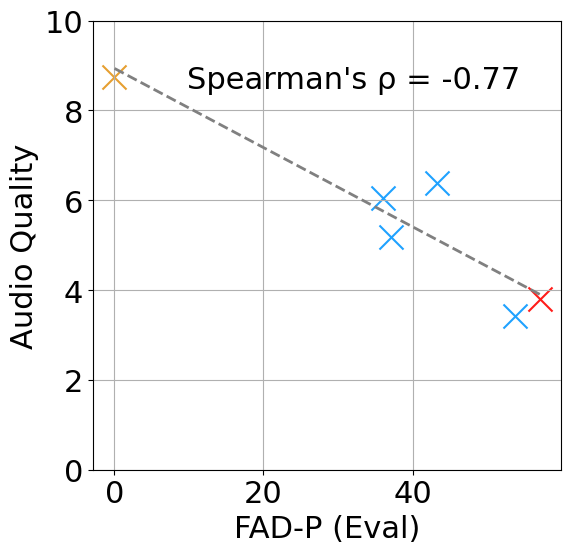}
        \subcaption{FAD-P on Evaluation set vs Audio Quality}
        \label{fig:eval-mos-aq}
    \end{subfigure}
    \caption{Correlation between FAD scores on evaluation set and other indicators, computed on the 4 submitted systems and the baseline system.}
    \label{fig:results}
\end{figure*}

\subsection{Objective Evaluation}
We employed the Fréchet Audio Distance (FAD)~\cite{fad} with PANN-Wavegram-Logmel~\cite{panns} embeddings as our primary objective metric. The embedding was chosen to maximize the correlation between the FAD score and the human perception \cite{fad_audio}. The FAD computation is defined as:

\begin{equation}
    \text{FAD}(r,g) = \|\mu_r-\mu_g\|^2 + \text{Tr}(\Sigma_r+\Sigma_g-2\sqrt{\Sigma_r\Sigma_g})
\end{equation}

In this formulation, $r$ and $g$ represent the reference and generated audio sets respectively, with their corresponding mean feature vectors $\mu_r, \mu_g$ and covariance matrices $\Sigma_r, \Sigma_g$. This metric effectively captures both the statistical similarities between the generated and reference audio distributions and their feature relationships.
We provided an official evaluation software.\footnote{
\scriptsize{\url{https://github.com/DCASE2024-Task7-Sound-Scene-Synthesis/fadtk}}
}

\subsection{Subjective Evaluation}
Our subjective evaluation framework assessed three key aspects of the generated audio, each rated on a 0-10 scale. The Foreground Fit (FF) measured the accuracy of the primary sound source, while Background Fit (BF) evaluated the appropriateness of the ambient sound. Overall Audio Quality (AQ) captured the perceptual quality of the generated audio. We computed a weighted final perceptual score that emphasized foreground accuracy while balancing background fit and audio quality:

\begin{equation}
    \text{Perceptual Score} = \frac{2\text{FF} + \text{BF} + \text{AQ}}{4}
\end{equation}

\section{Results}

\subsection{System Performance}
Table \ref{tab:results} summarizes the evaluation results.
The evaluation process encompassed four submitted systems~\cite{Sun_Samsung,Chung_KT,Yi_Surrey,Verma_IITM} assessed by a panel of 14 expert raters comprising four participants from competing teams and ten challenge organizers. We systematically evaluated 24 carefully selected evaluation captions, ensuring equal representation with four samples per foreground category. These evaluation prompts were selected ahead of time. Raters and data analyzers were blind to the identities of the system creators. Self-ratings were removed. This balanced approach allowed for comprehensive assessment across all sound categories while maintaining manageable evaluation time constraints. The inter-rater agreement was very high (Cronbach’s alpha = 0.959), validating our careful rating methodology.
For detailed system characteristics and performance scores, refer to our result webpage.\footnote{\scriptsize{\url{https://dcase.community/challenge2024/task-sound-scene-synthesis-results}}}

\begin{table}[h!]
\centering
\resizebox{\columnwidth}{!}{%
\begin{tabular}{@{}lcccc@{}}
\toprule
\textbf{Submission Code} & \textbf{Official Rank} & \textbf{Average Score} & \textbf{FAD (PANNs)} & \textbf{ML Method} \\ \midrule
Sound Designer (Ref.) & - & 8.793 & 0      & -                \\
Sun\_Samsung~\cite{Sun_Samsung}          & 1 & 5.832 & 35.985 & Latent Diffusion \\
Chung\_KT~\cite{Chung_KT}             & 2 & 4.966 & 37.092 & GAN              \\
Yi\_Surrey~\cite{Yi_Surrey}            & 3 & 4.748 & 43.304 & Latent Diffusion \\
DCASE2024\_baseline~\cite{liu2023audioldm}   & - & 3.287 & 57.061 & Latent Diffusion \\
Verma\_IITMandi~\cite{Verma_IITM}       & 4 & 2.523 & 53.728 & Latent Diffusion \\ \bottomrule
\end{tabular}%
}
\caption{Challenge results: official ranks and evaluation scores.}
\label{tab:results}
\end{table}

\subsection{Analysis}
Our comprehensive analysis revealed several key findings from the results shown in Table \ref{tab:results} and Figure \ref{fig:results}. Most notably, we observed a substantial 36\% performance gap between the sound engineer reference and the best submitted system, indicating significant room for improvement in synthetic audio quality. We found a strong correlation between objective FAD scores and subjective metrics, with a correlation coefficient of 0.94 for foreground fit, 0.94 for background fit, and 0.77 for audio quality, meaning that most, but not all, of the variance is shared between our objective and subjective measures. However, this result should be considered as a weak evidence for the effectiveness of FAD as a perceptual measure, since the number of data points is small. 

For more detailed analysis, please refer to another paper of ours, Lee~et.~al.~\cite{lee2024challenge}, since we focus on reporting the challenge result in this paper.

\subsection{Participation Analysis}
The challenge saw a notable decrease in participation compared to the previous year~\cite{foley2023}, with 4 submissions in 2024 versus 32 in 2023. This reduction may be attributed to several factors: the broader task scope that favored teams with access to large pre-existing models, the removal of training dataset constraints that previously encouraged wider participation, changes in task naming and framing that may have affected its appeal, and the introduction of more complex evaluation requirements that increased the entry barrier for potential participants.

\section{Discontinuation of the Task}
It is worth mentioning why the organizers decided not to continue the DCASE challenge in 2025 despite the successful challenges in 2023 and 2024. 

First, the generative aspect of organizing this challenge has been costly and labor intensive. In this year's challenge, it took (a) about 40 hours to create and refine the evaluation set of sounds created by a sound designer, (b) about 40 hrs to debug and run the code for 4 participants and (c) about 40 hours to conduct the perceptual evaluation, including organizer time to create and analyze the rater survey plus rater time (raters listed in the acknowledgments). Those efforts were specific to the generative aspect of the task and were therefore in addition to our other efforts that are a typical part of organizing a classification task, such as collating audio samples and making a baseline system. Furthermore, since DCASE results are announced shortly after participants' systems are submitted, there was very little time for the extra steps of generating sounds from each submitted system and having raters perceptually evaluate them. In addition, there were financial costs for system evaluation, such as running participant-supplied code in Colab. 

Second, the nature of generative audio has been evolving over the past two years.  When first organizing this challenge, our problem formulation was specific so that we could not only evaluate systems, but also explain the results and compare the sound qualities from year to year. However, this meant the scope of our evaluation was narrower than those of many existing models. Currently, the overall scope of the academic community is expanding and diversifying while the topic of each paper becomes more specialized, e.g., by focusing on temporal alignment~\cite{tfoley,mambafoley,multi_con_diff,picoaudio}, high-fidelity audio, foundational models for general sound generation~\cite{audioldm2,auffusion,ezaudio,fugatto,stableaudio,tango-llm}, video-as-an-input~\cite{syncfusion,sonicvisionlm,video-foley,rewas,moviegen,multifoley,kushwaha2024vintage,frieren,maskvat,v-aura,ssv2a,mmaudio}, etc. 

\section{Conclusion}
The DCASE 2024 Challenge Task 7 has provided valuable insights into the current state of sound scene synthesis while highlighting several crucial areas for future development. While the submitted systems demonstrated promising capabilities, the significant gap between synthetic and reference audio quality indicates substantial room for improvement. Future developments should focus on implementing more complex caption structures, supporting multiple foreground sounds, and enhancing synthesis techniques to reduce the quality gap. Additionally, the development of more sophisticated evaluation metrics for sound scene coherence will be crucial for advancing the field.

Looking ahead, we envision several key directions for the challenge's evolution. These include expanding the complexity of supported scenes, improving the evaluation framework to better capture nuanced aspects of generated audio, and developing more efficient training approaches that enable broader participation. We believe this standardized evaluation framework will continue to drive progress in sound scene synthesis, ultimately leading to more realistic and versatile audio generation systems.

\section{Acknowledgements}
We thank all the raters who did the subjective evaluation: Xie ZhiDong, Li XinYu, Liu HaiCheng, Zou XiaoYan, Sun Yu, Hae Chun Chung, Jae Hoon Jung, Yi Yuan, Haohe Liu, Xubo Liu, Mark D. Plumbley, Wenwu Wang, Sagnik Ghosh, Gaurav Verma, Siddharath Narayan Shakya, Shubham Sharma, Shivesh Singh, Urszula Oszczapinska, Paige Brady, Angjelica Ferguson, Sripathi Sridhar, and organizers without conflicts  -- Modan Tailleur, Keunwoo Choi, Keisuke Imoto, Brian Mcfee, Yuki Okamoto, Laurie M.~Heller. We thank Justine Sullivan and Anjelica Ferguson for help with creating and analyzing the perceptual tests. We thank CMU Psychology for underwriting the cost of the survey platform for ratings, Mathieu Lagrange for helping with the cost of computing resources, and Applied Media Information Laboratory, Doshisha University for helping with the cost of the Slack channel for organizer communication.

\bibliographystyle{IEEEtran}
\bibliography{refs}

\end{sloppy}
\end{document}